\documentclass[pdflatex,sn-nature]{sn-jnl}

\usepackage{graphicx}
\setcitestyle{super,open={},close={}}

\usepackage{amsmath,amssymb}

\usepackage{booktabs}

\usepackage{textcomp}

\begin{document}

\setcounter{secnumdepth}{0}

\title{dMRI-F: A Foundation Model for Generalizable Diffusion MRI Analysis from Sparse Acquisitions}

\author{\fnm{Xi} \sur{Zhu}\textsuperscript{\smash{1}}}

\author{\fnm{Yijie} \sur{Li}\textsuperscript{\smash{1}}}

\author{\fnm{Xiaofan} \sur{Wang}\textsuperscript{\smash{1}}}

\author{\fnm{Wei} \sur{Zhang}\textsuperscript{\smash{1}}}

\author{\fnm{Huafu} \sur{Chen}\textsuperscript{\smash{1}}}

\author{\fnm{Ce} \sur{Zhu}\textsuperscript{\smash{1}}}

\author*{\fnm{Fan} \sur{Zhang}\textsuperscript{\smash{1,*}}}
\email{fan.zhang@uestc.edu.cn}


\affil[1]{\orgname{University of Electronic Science and Technology of China}}

\abstract{Diffusion MRI (dMRI) probes water diffusion within the brain, providing unique information about tissue microstructure and white matter connectivity. High-quality dMRI enables advanced and accurate analyses but demands dense acquisitions, resulting in long scan times and hardware constraints that are often infeasible. Further, dMRI requires dedicated computational methods to support diverse analysis tasks (e.g., parcellation, registration, and synthesis) for both neuroscientific and clinical applications. Here, we present dMRI-F, a foundation model that unifies dMRI data enhancement and downstream tasks across diverse acquisitions and populations, while requiring only minimal acquisition input. dMRI-F is pre-trained with a q-space masked reconstruction task: predicting densely sampled multi-shell data from only six volumes. It learns generalizable latent feature representations of dMRI data, which serve as a core foundation for adapting to diverse downstream tasks, including brain parcellation, brain registration, and cross-modal synthesis. Across nine independently acquired datasets, we demonstrate that dMRI-F outperforms state-of-the-art methods with robust generalization across acquisitions, populations, and tasks. Our results demonstrate dMRI-F's versatility as a general-purpose framework that enables diverse dMRI applications from low-cost, sparse acquisitions.}

\maketitle


Diffusion magnetic resonance imaging (dMRI) is an advanced, non-invasive imaging technique that can uniquely measure the random motion of water molecules within the brain, enabling \emph{in vivo} characterization of tissue microstructure\cite{ref1} and structural connectivity\cite{ref2}. In dMRI, the acquired data, known as diffusion-weighted images (DWIs), are collected by sampling q-space, where each sampled location corresponds to a DWI volume acquired with a specific gradient direction (i.e., b-vector) and strength (i.e., b-value). From the DWIs, biophysical models can be fit to derive microstructural measures (e.g., fractional anisotropy, FA)\cite{ref3}, and fiber tractography can be performed to map the brain's structural connectivity\cite{ref4}. Leveraging these properties, dMRI has found broad utility in both neuroscientific research and clinical practice, with applications spanning brain development\cite{ref5}, aging\cite{ref6}, neuroplasticity\cite{ref7}, and cognitive function\cite{ref8}, as well as lesion characterization\cite{ref9}, disease monitoring\cite{ref10}, and neurosurgical planning\cite{ref11}.

A key demand in dMRI is acquiring high-quality DWI data, combining dense q-space sampling, high spatial resolution, and robust signal fidelity, to improve the accuracy of modeling subtle microstructural features and complex fiber configurations. However, acquiring such high-quality data entails prolonged scan times and substantial hardware demands. The Human Connectome Project Young Adult (HCP-YA) project\cite{ref12} exemplifies this trade-off: its dMRI protocol achieves high 1.25\,mm isotropic resolution, dense 270 diffusion directions across three different shells, and multiple phase-encoding acquisitions for artifact correction, but requires customized scanners and nearly one hour of scanning time per subject. While the HCP-YA data have proven immensely valuable to the neuroimaging and neuroscientific community\cite{ref13,ref14,ref15}, this level of quality remains impractical for most routine studies, which typically acquire single-shell acquisitions with \textasciitilde{}2\,mm isotropic resolution and far fewer gradient directions---often as few as 6, the minimum required for conventional diffusion tensor modeling\cite{ref16}. Consequently, practical dMRI remains quality-limited, necessitating efficient DWI enhancement methods that operate under routine acquisition constraints.

Deep learning has emerged as a promising approach for enhancing dMRI quality by inferring high-quality DWI data from sparse acquisitions. Existing methods generally fall into three major categories based on their primary objectives. The first focuses on angular super-resolution, where densely sampled diffusion signals encompassing a large number of gradient directions across multiple b-values are inferred from sparsely acquired q-space data\cite{ref17,ref18,ref19}. The second category emphasizes spatial super-resolution, aiming at enhancing the spatial resolution of DWI data by generating fine-grained structural details from low-resolution scans\cite{ref20,ref21,ref22}. The third category addresses denoising, with the goal of reducing imaging noise to improve the fidelity of diffusion signals\cite{ref23,ref24}. Overall, these methods aim to capture the complex relationships inherent to the high-dimensional dMRI signal space---spanning both spatial (3D) and diffusion-encoding (q-space) dimensions---to enable prediction of enhanced signal data. However, despite their success, most existing methods process the data as independent DWI slices, limiting their ability to exploit its full high-dimensional spatial continuity. Furthermore, these methods are typically tailored to specific enhancement objectives and acquisition protocols, leaving a strong need for a unified framework for broad applicability.

Beyond data-quality enhancement, dMRI also demands accurate and efficient methods for various computational tasks. Unlike traditional anatomical MRI (e.g., T1w), dMRI uniquely characterizes water diffusion using multi-directional, multi-shell acquisitions. The resulting large number of image volumes (DWIs) creates substantial computational challenges and requires dedicated analytical methods. For example, in brain parcellation, recent studies have sought to perform the task directly on DWIs by extracting tissue-informative microstructural contrasts, thereby bypassing the inter-modality registration required by conventional T1w-based approaches\cite{ref25,ref26,ref27}. Similarly, in brain registration, an emerging trend is to incorporate multiple complementary dMRI-derived maps that provide distinct tissue and fiber orientation information to guide image alignment\cite{ref28,ref29,ref30}. Furthermore, cross-modal synthesis methods generate T1w images directly from DWIs when native anatomical images are unavailable or cannot be reliably aligned with diffusion data\cite{ref31,ref32}. Across these distinct tasks, the common goal is to extract anatomically informative features directly from dMRI. Yet, the pursuit of feature representations that are broadly applicable across various tasks remains underexplored.

Another demand in dMRI is generalizable methods across diverse acquisitions and populations. dMRI is known to be highly sensitive to acquisition protocols, scanner types, and parameter settings, which often undermines a method's generalizability across studies\cite{ref33,ref34}. In related fields in neuroimaging, one common strategy to improve generalizability is to train on large-scale pooled datasets that increase data variability\cite{ref35,ref36}---though this typically requires substantial data curation efforts and high computational costs. In dMRI, an emerging alternative is to leverage high-quality reference datasets (e.g., HCP-YA) that capture a broad distribution of diffusion properties, with the aim of generalizing to clinically typical acquisitions and populations. For example, the strategy of training on HCP-YA data and applying to other datasets has proven successful across various tasks, including brain segmentation\cite{ref37} and tract parcellation\cite{ref38,ref39,ref40}, with demonstrated generalizability across datasets spanning diverse genders, ages, and health conditions. Central to this strategy is the extraction of informative contrasts that provide compact yet discriminative representations---essential for achieving robust generalizability.

Here, we introduce dMRI-F, a foundation model for dMRI that jointly performs DWI enhancement and learns generalizable latent feature representations, enabling diverse downstream computational tasks across varying acquisition protocols and populations. dMRI-F employs a self-supervised learning (SSL) framework, pre-trained through a q-space masked-reconstruction task that predicts densely sampled multi-shell DWIs from sparse single-shell inputs (\hyperlink{dmrif-figure1}{Fig. 1a}). Architecturally, the model comprises an anatomy-aware encoder that learns latent features reflecting the underlying anatomical structures from input volumes, and a coordinate-conditioned decoder that predicts unseen volumes by jointly modeling these features with DWI spatial and q-space coordinates (\hyperlink{dmrif-figure1}{Fig. 1b}). By generating enhanced, densely sampled data, dMRI-F enables advanced analysis of tissue microstructure and white-matter connectivity from sparse inputs of only six volumes (\hyperlink{dmrif-figure1}{Fig. 1c}). Furthermore, by fine-tuning the learned latent features, dMRI-F enables generalizable adaptation to diverse downstream tasks, including brain parcellation, registration, and cross-modal synthesis (\hyperlink{dmrif-figure1}{Fig. 1d}). Using high-quality data from only 200 HCP-YA subjects for training, dMRI-F achieves robust generalization for DWI enhancement and downstream tasks across various acquisitions and populations (1,882 unseen subjects from the high-quality HCP-YA test set and eight independent test datasets with clinically typical acquisitions) (\hyperlink{dmrif-figure1}{Fig. 1e}). Together, our results demonstrate dMRI-F as a versatile, general-purpose computational framework for dMRI, offering a solid foundation for diverse neuroscientific and clinical applications with minimal acquisition requirements.

\begin{figure}[!t]
\begin{center}
\hypertarget{dmrif-figure1}{}%
\includegraphics[width=\linewidth,height=0.62\textheight,keepaspectratio]{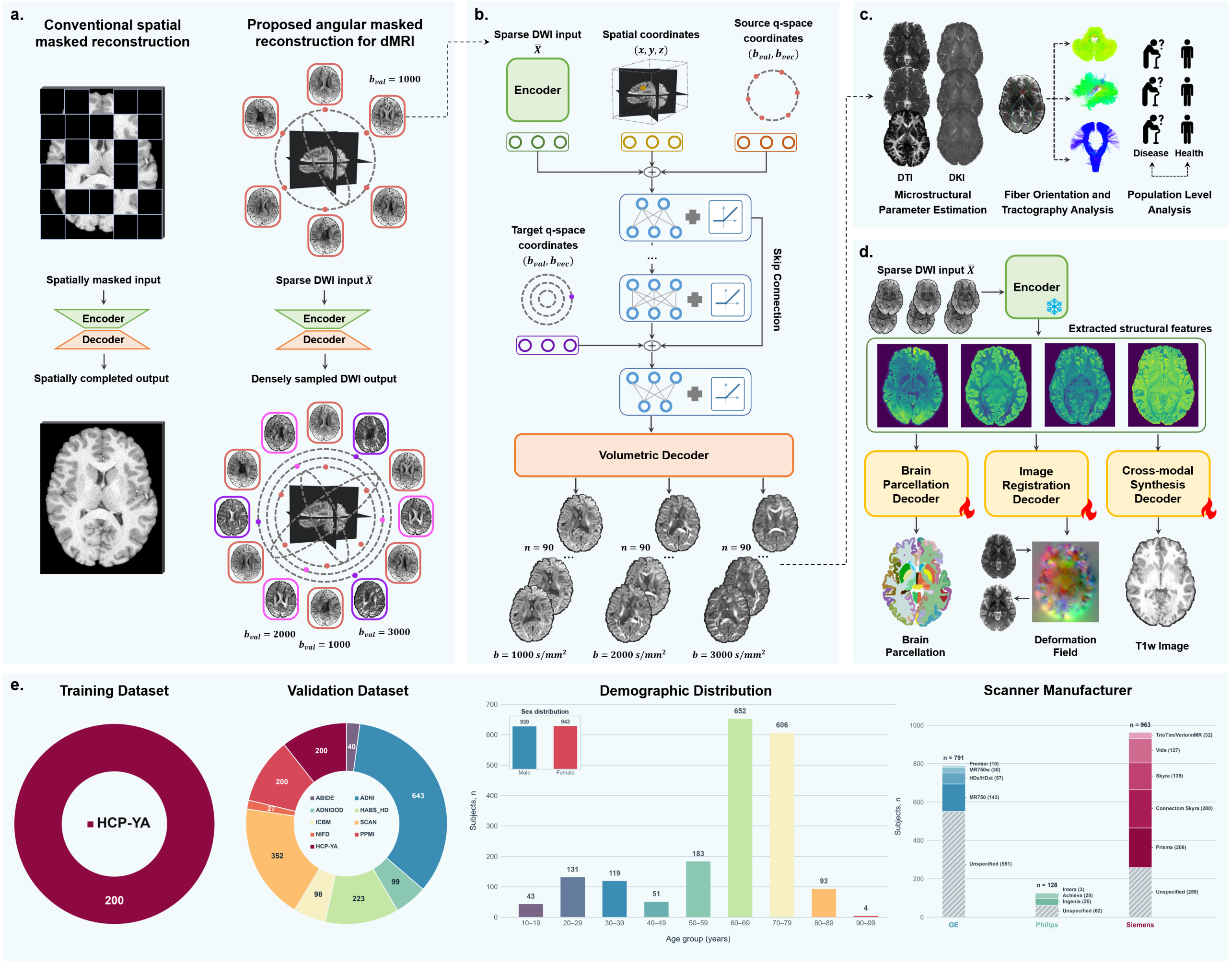}
\end{center}

\noindent\textbf{\textbf{Fig. 1 | Overview of dMRI-F and the study design.}}\par

\textbf{a,} \textbf{Pretraining Task}: Extending the principle of conventional spatial masked reconstruction, dMRI-F performs q-space masked reconstruction to predict 270 densely sampled multi-shell DWIs across b = 1,000, 2,000, and 3,000\,s/mm\textsuperscript{2} using only six single-shell volumes at b = 1,000\,s/mm\textsuperscript{2} as input. \textbf{b,} \textbf{Network Architecture}: dMRI-F includes an anatomy-aware encoder that extracts latent feature representations describing the underlying anatomical structures from the sparse DWI inputs, and a Neural Radiance Fields (NeRF)-inspired coordinate-conditioned decoder that models the relationships between the latent features and the corresponding spatial and q-space coordinates to predict the target DWI signals. \textbf{c, DWI Enhancement: }The predicted densely sampled DWI data enable advanced analysis of white-matter connectivity and tissue microstructure---analyses that are not feasible from the original six-volume input alone. \textbf{d,} \textbf{Downstream Task Adaption}: The learned latent feature representations support generalized adaptation to diverse downstream tasks, including brain parcellation, brain registration, and cross-modal synthesis. \textbf{e}\textbf{,} \textbf{Training and Evaluation Data}: dMRI-F was pre-trained on high-quality dMRI data from 200 HCP-YA subjects and evaluated on 1,882 unseen subjects, comprising the HCP-YA test set and eight independent datasets with clinically typical acquisitions.
\end{figure}

\section{Results}

\subsection{Overview}

We demonstrate dMRI-F from the following aspects. First, we present dMRI-F's advantages in DWI enhancement to predict densely sampled data from sparse, low-b-value inputs. These include: (1) a benchmark comparison with state-of-the-art (SOTA) methods on high-quality HCP-YA data; (2) a generalization assessment across eight independent test datasets from diverse acquisition protocols and populations; and (3) a fidelity evaluation of the enhanced DWIs for preserving white matter connectivity and population tissue microstructure. Next, we showcase dMRI-F's capability for downstream computational tasks by fine-tuning task-specific decoders on the learned latent representations. These include: (1) brain parcellation, which subdivides DWIs into anatomically segregated brain regions; (2) brain registration, which aligns DWI volumes across different subjects; and (3) cross-modal synthesis, which generates anatomical T1w images from DWIs. For each task, we benchmarked dMRI-F against representative SOTA methods and also assessed generalizability on both the HCP-YA and the test datasets with clinically typical acquisitions. Finally, we assess our method design, including: (1) an ablation study evaluating the network architecture and major parameters (see ``\hyperlink{dmrif-extended-table1}{Extended Data Table 1}''); (2) an analysis of the effect of training subject size (see ``\hyperlink{dmrif-extended-figure1}{Extended Data Fig. 1}''); and (3) an evaluation of model performance under different numbers of input diffusion directions (see ``\hyperlink{dmrif-extended-figure2}{Extended Data Fig. 2}'').

\subsection{dMRI-F Robustly Predicts Dense DWI Data from Sparse Inputs Across Acquisitions and Populations}

\subsubsection{Superior DWI Enhancement Performance \textbf{on }Benchmark \textbf{HCP-YA Data}}

\begin{figure}[!htbp]
\begin{center}
\hypertarget{dmrif-figure2}{}%
\includegraphics[width=\linewidth,height=0.62\textheight,keepaspectratio]{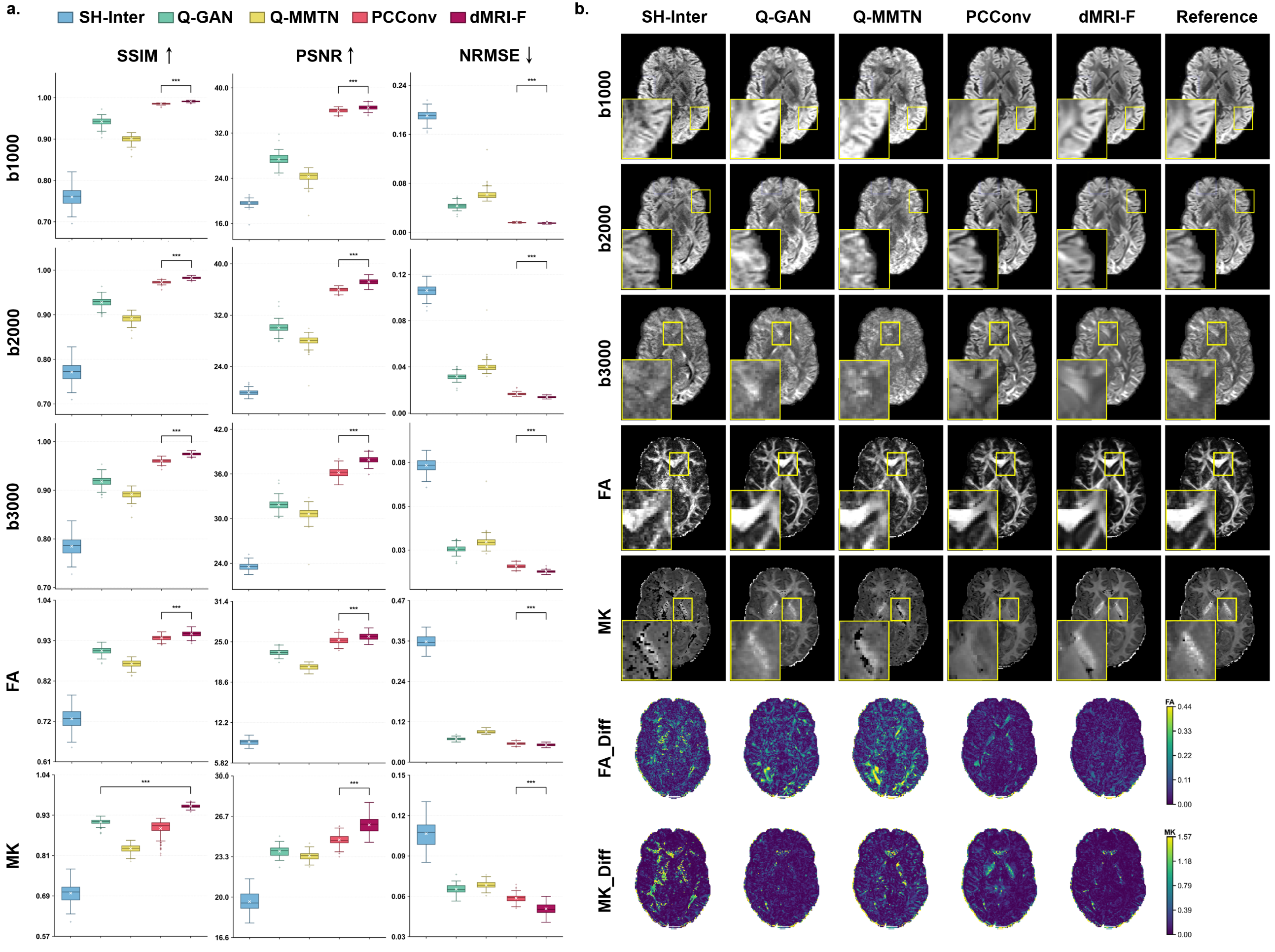}
\end{center}

\noindent\textbf{\textbf{Fig. 2 | dMRI-F achieves superior data enhancement performance in both DWI signal prediction and microstructural measure estimation in the benchmark HCP-YA data.}}\par

\textbf{a}\textbf{, }Quantitative comparisons between dMRI-F and the SOTA DWI enhancement methods (SH-Inter, Q-GAN, Q-MMTN, and PCConv) at both the DWI signal (at b = 1,000, 2,000, and 3,000\,s/mm\textsuperscript{2}) and derived microstructure measure (FA and MK) levels. \textbf{b,} Visualization of example DWIs, FA maps, and MK maps derived from each method, together with the corresponding reference images from the full DWI data. Yellow boxes and zoomed views highlight example local regions. FA\_Diff and MK\_Diff show absolute errors of each method relative to the reference. \textit{Note:} For all box plots, the center line denotes the median, the box spans the interquartile range, and the whiskers extend to 1.5 times the interquartile range. White diagonal crosses denote the mean, whereas plus signs mark outliers beyond the whiskers. Significance is shown for the comparisons marked by brackets: *\emph{p < 0.05}, **\emph{p < 0.01}, and ***\emph{p < 0.001}.
\end{figure}

We first evaluated dMRI-F's ability on our pretraining DWI enhancement task, i.e., predict densely sampled multi-shell data (270 DWI volumes across b = 1,000, 2,000, and 3,000\,s/mm\textsuperscript{2}) from sparse, low-b-value input (6 volumes at b = 1,000\,s/mm\textsuperscript{2}). This was conducted on a benchmark test set of 200 HCP-YA subjects. We compared dMRI-F with four SOTA methods: (1) SH-Inter\cite{ref41}, (2) Q-GAN\cite{ref18}, (3) Q-MMTN\cite{ref17}, and (4) PCConv\cite{ref19}. Briefly, SH-Inter performs interpolation in the spherical harmonics domain, Q-GAN and Q-MMTN learn sparse-to-dense DWI mappings with additional T1w and T2w inputs, whereas PCConv learns a direct mapping from only DWI data (see Supplementary Table 1 for details). Performance was evaluated against the full DWI data as reference, at two levels: (1) DWI signals at each b value, and (2) microstructure measures (including fractional anisotropy; FA, mean kurtosis; MK, etc.) derived for diffusion kurtosis (DK) modeling\cite{ref42}. We used three metrics: Normalized Root Mean Square Error (NRMSE) for voxel-wise error, Peak Signal-to-Noise Ratio (PSNR) for peak error derived from MSE, and Structural Similarity Index Measure (SSIM) for contrast and structural similarity.

dMRI-F outperformed each SOTA method at both evaluation levels (\hyperlink{dmrif-figure2}{Fig. 2a}). For the DWI signals, dMRI-F achieved statistically significant improvements over PCConv, the strongest compared method, with higher PSNR and SSIM and lower NRMSE. This gain was also evident at the derived microstructure measures, where dMRI-F consistently outperformed all competing methods in the FA and MK maps (see Supplementary Table 2 for results of additional maps) and each metric. Notably, dMRI-F maintained its advantage across all three diffusion shells, with particularly strong performance at the higher b-value shells. This more accurate recovery of high-b-value signals was reflected in the pronounced gains observed for the kurtosis-related MK maps. Visual comparisons showed that the predicted DWIs from dMRI-F most closely matched the reference across all three shells, with better-preserved tissue boundaries and structural details (\hyperlink{dmrif-figure2}{Fig. 2b} and Supplementary Fig. 1). The FA and MK maps derived from dMRI-F likewise exhibited clearer tissue contrast and fewer local artifacts, together with visibly lower residual errors than those of the compared methods.

\subsubsection{Robust Generalization for DWI Enhancement Across Independently Acquired Test Datasets}

To assess whether dMRI-F generalizes beyond the HCP-YA training data under clinically realistic conditions, we evaluated the trained DWI enhancement model on eight test datasets comprising 1,682 subjects (see \hyperlink{dmrif-methods-testing-data}{Methods} and Supplementary Table 3 for demographic and acquisition details). These cohorts spanned a wide age range from 10 to 95 years, included both healthy and multiple disease conditions, and were acquired across multiple scanner manufacturers and protocols covering diffusion shells at 1,000, 1,500, or 2,000\,s/mm\textsuperscript{2}. In contrast to the high-quality HCP-YA training set, these test datasets had lower spatial resolution, higher noise, and substantial cross-site image-quality variation, providing a stringent test of dMRI-F's generalizability across diverse data conditions. We compared dMRI-F against PCConv---the top-performing method in the preceding benchmark evaluation---under identical conditions for predicting the densely sampled DWI data from sparse inputs of six volumes.

Across the eight test datasets, dMRI-F significantly outperformed PCConv across all quantitative metrics, at both the DWI signal and derived microstructural measure levels (\hyperlink{dmrif-figure3}{Fig. 3a}). Notably, the improvement was particularly evident on the ABIDE dataset, which was acquired at b = 1,500\,s/mm\textsuperscript{2}, a shell not observed during model training on HCP-YA. Qualitatively, despite substantial variation in input quality, the DWIs and FA images derived from dMRI-F were visually highly consistent with the reference data. The zoomed images reveal that dMRI-F better recovered tissue boundaries and local anatomical patterns, whereas PCConv often produced smoother predictions with less distinct anatomical contrast (\hyperlink{dmrif-figure3}{Fig. 3b}). Collectively, these results showed that dMRI-F generalized robustly across heterogeneous imaging conditions and populations, with the ability to extrapolate to diffusion shells beyond those encountered during training.

\begin{figure}[!htbp]
\begin{center}
\hypertarget{dmrif-figure3}{}%
\makebox[\linewidth][c]{\includegraphics[width=1.06\linewidth,height=0.75\textheight,keepaspectratio]{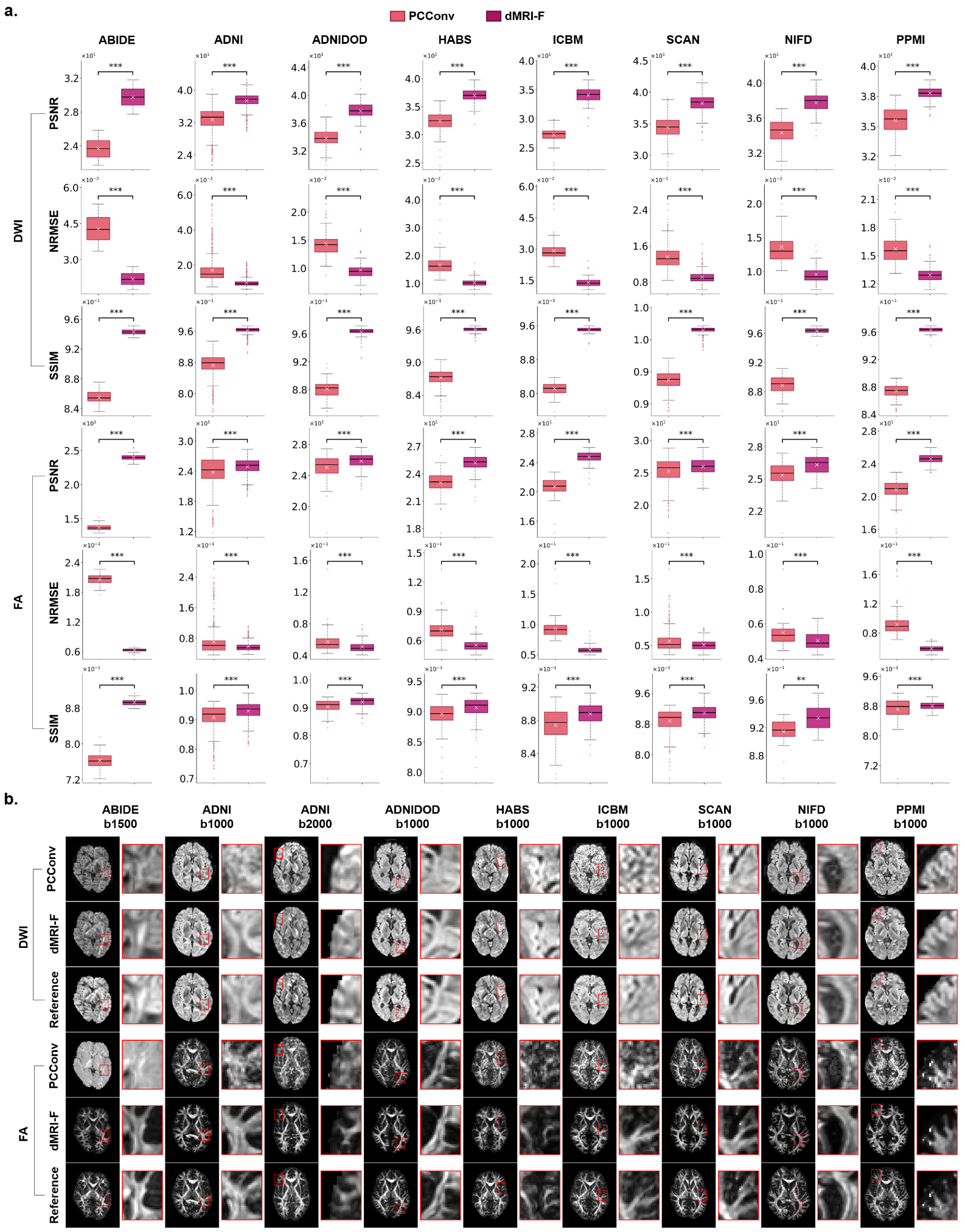}}
\end{center}

\noindent\textbf{\textbf{Fig. 3 | dMRI-F enables robust generalizations for DWI enhancement across acquisition protocols and populations.}}\par

\textbf{a,} Quantitative comparison of dMRI-F and PCConv for DWI signal prediction and microstructural estimation across the multiple testing datasets with clinically typical acquisitions. \textbf{b}\textbf{,} Visualization of the predicted DWIs and corresponding derived FA maps in each test dataset. Separate ADNI examples are shown for the b = 1,000 and 2,000\,s/mm\textsuperscript{2} shells, and ABIDE represents an unseen b = 1,500\,s/mm\textsuperscript{2} acquisition. Red boxes and enlarged views highlight local tissue boundaries and anatomical details. \textit{Note:} For all box plots, the center line denotes the median, the box spans the interquartile range, and the whiskers extend to 1.5 times the interquartile range. White diagonal crosses denote the mean, whereas plus signs mark outliers beyond the whiskers. Significance is shown for the comparisons marked by brackets: *\emph{p < 0.05}, **\emph{p < 0.01}, and ***\emph{p < 0.001}.
\end{figure}

\subsubsection{High Fidelity of Predicted Data for Preserving White Matter Connectivity and Population Tissue Microstructure}

We performed two post hoc analyses---fiber tractography and population-level microstructure comparison---to assess whether the enhanced DWI data preserve information essential for downstream applications.

For the tractography analysis, we processed the nine evaluation datasets (HCP-YA and eight test datasets) through a standard pipeline: fiber orientation distribution (FOD) estimation and whole-brain tractography (using MRtrix\cite{ref43}), followed by automated atlas-based tracts identification (using WhiteMatterAnalysis with the ORG atlas\cite{ref15,ref44}). This was applied to the predicted DWI data from our dMRI-F method and PCConv, as well as the reference full DWI data. Results showed that dMRI-F achieved significantly more accurate FOD estimation across all datasets (measured using the angular correlation coefficient), producing FOD maps visually similar to the reference (\hyperlink{dmrif-figure4}{Fig. 4a}). Visual comparisons of the derived fiber tracts (the corticospinal tract in \hyperlink{dmrif-figure4}{Fig. 4a} and additional tracts in Supplementary Fig. 2) further demonstrated that dMRI-F tractography more closely resembled the reference, whereas PCConv frequently yielded sparser trajectories and, in some cases, failed to reconstruct the fiber pathways.

For the microstructure analysis, building on the experiments demonstrating that the derived microstructural estimates were highly consistent with those obtained from the full DWI (see \hyperlink{dmrif-figure3}{Fig. 3} in the above experiment), we further conducted two population-level comparisons: a sex comparison (female vs. male) and a disease comparison (Alzheimer's disease, AD vs. normal controls, NC). For the sex comparison, we included 20 age-matched female--male pairs from the HCP-YA testing cohort, computed the mean fractional anisotropy (FA) across the whole white-matter skeleton---an interpretable marker of sex-related white-matter organization\cite{ref45}---and compared it between groups using a two-sided t-test. For the disease comparison, we included 20 age-matched AD--NC pairs from the ADNI\cite{ref46} dataset and compared mean skeleton MD (a highly sensitive measure for detecting AD-related white-matter abnormalities\cite{ref47}) between the two groups. We observed significantly higher FA in females in the sex comparison and significantly higher MD in the AD group in the disease comparison, with the detected group differences being highly consistent between the predicted and full-reference data (\hyperlink{dmrif-figure4}{Fig. 4b}).

Overall, these results suggest that dMRI-F-enhanced DWIs largely preserve fidelity for FOD estimation and enable anatomically reliable tractography across varied acquisition conditions. Moreover, the enhanced data appear to retain known population-level microstructure differences, consistent with those observed in the full reference data.

\begin{figure}[!htbp]
\begin{center}
\hypertarget{dmrif-figure4}{}%
\makebox[\linewidth][c]{\includegraphics[width=1.06\linewidth,height=0.75\textheight,keepaspectratio]{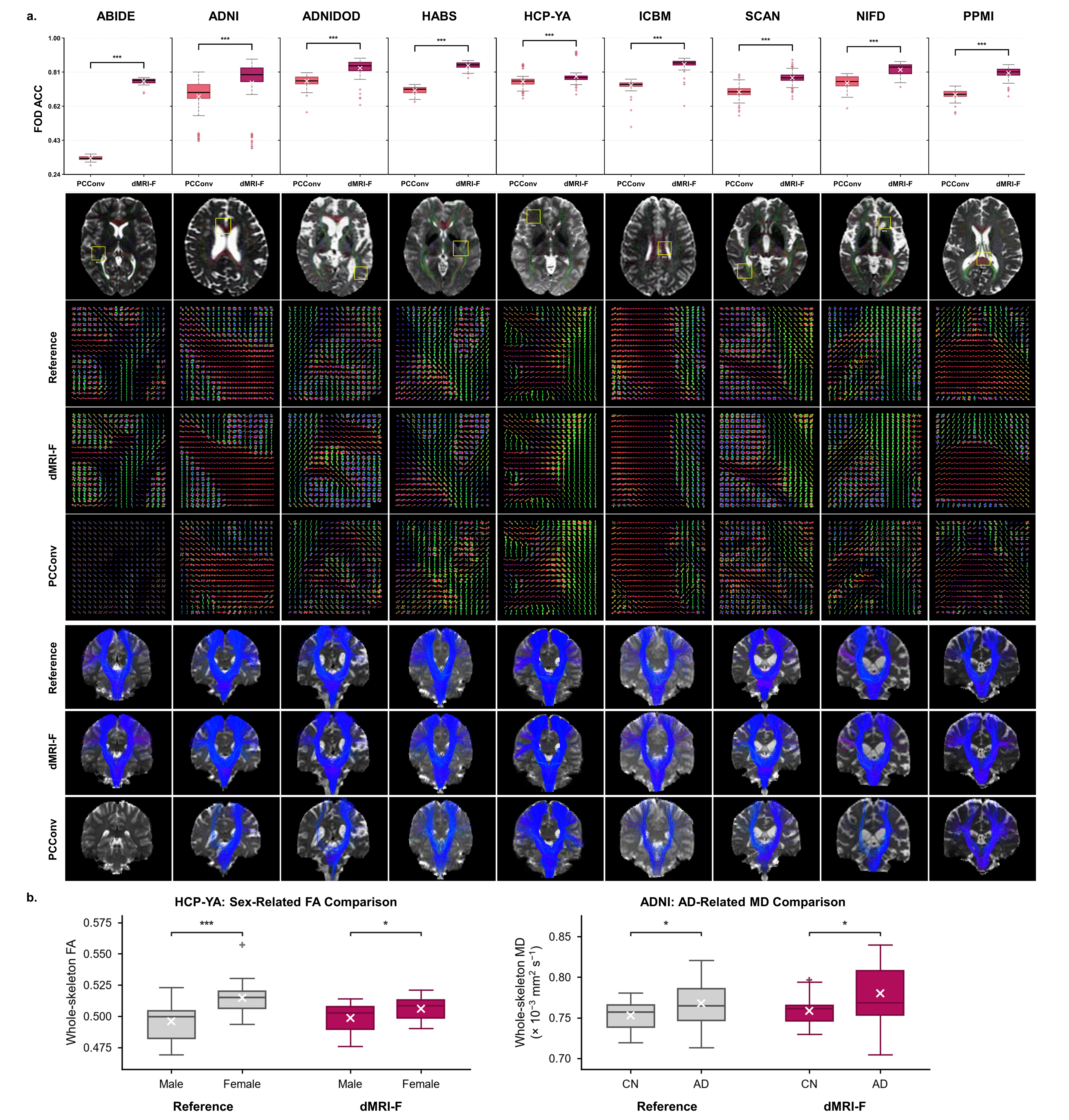}}
\end{center}

\noindent\textbf{\textbf{Fig. 4 |}\textbf{ dMRI-F's predicted DWIs preserve fiber-orientation information for tractography and maintain microstructure information for population-level comparisons. }}\par

\textbf{a,} Quantitative comparison of FOD estimation accuracy between dMRI-F and PCConv, with visualization of example FOD maps and derived fiber tracts from both methods alongside the reference data. \textbf{b,} Population-level microstructure analyses in HCP-YA for sex difference comparison (left) and ADNI for disease difference comparison (right). \textit{Note:} For all box plots, the center line denotes the median, the box spans the interquartile range, and the whiskers extend to 1.5 times the interquartile range. White diagonal crosses denote the mean, whereas plus signs mark outliers beyond the whiskers. Significance is shown for comparisons marked by brackets: *\emph{p < 0.05}, **\emph{p < 0.01}, and ***\emph{p < 0.001}.
\end{figure}

\subsection{dMRI-F Learned Feature Representations Enable Adaptation to Diverse Downstream Tasks}

\begin{figure}[!htbp]
\begin{center}
\hypertarget{dmrif-figure5}{}%
\includegraphics[width=\linewidth,height=0.62\textheight,keepaspectratio]{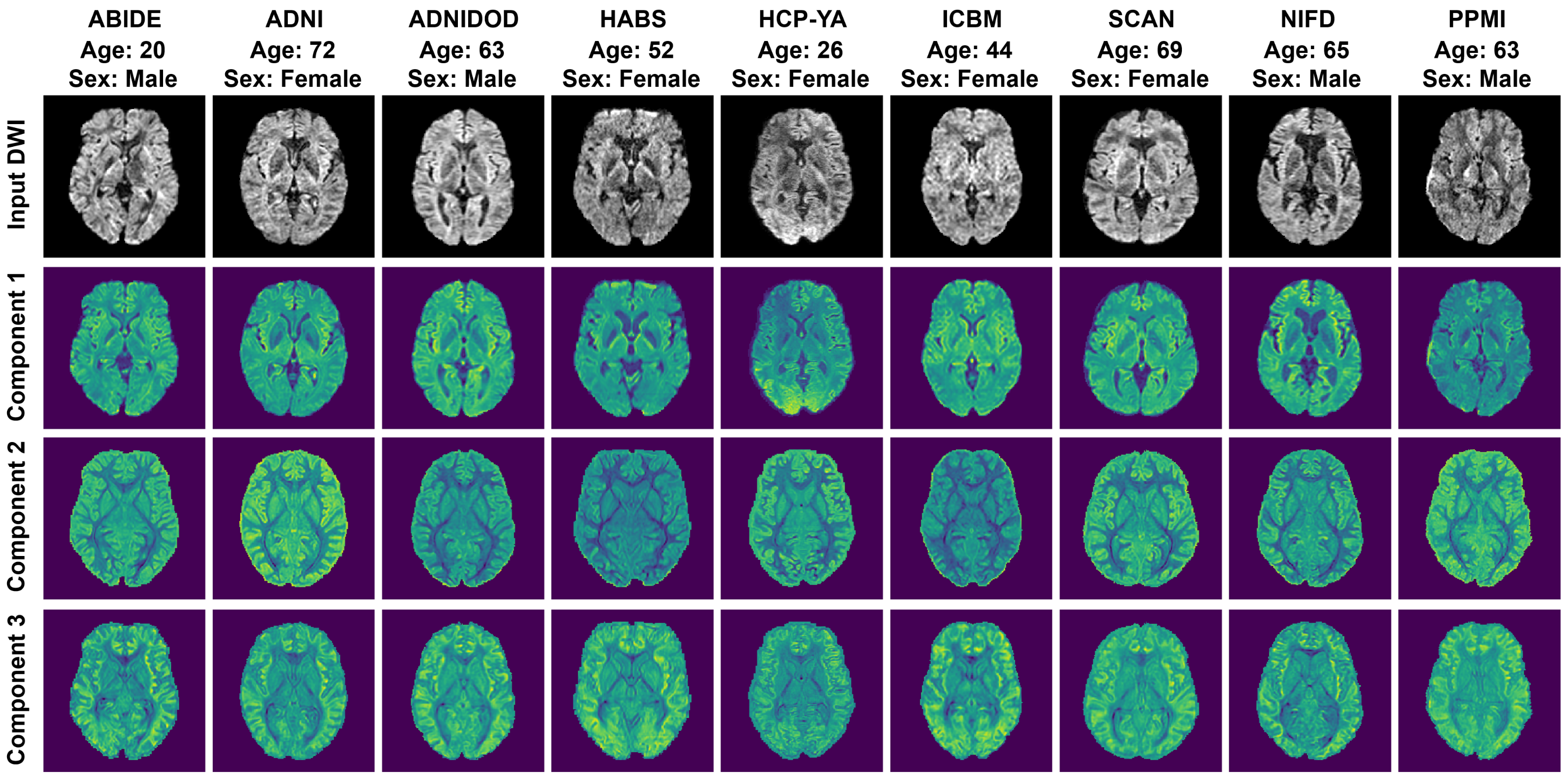}
\end{center}

\noindent\textbf{\textbf{Fig. 5 | dMRI-F learns consistent }\textbf{feature representations}\textbf{ across datasets. }}\par

Visualization of feature maps obtained from the pretrained dMRI-F encoder for example subjects from each evaluation dataset. The top row shows one DWI volume from each subject's input for reference. For visualization purposes, the first three principal components obtained by applying principal component analysis (PCA) to the features are shown for each subject. Despite substantial differences in acquisition protocols, image quality and study populations, each component consistently emphasizes similar tissue contrasts.
\end{figure}

The dMRI-F encoder enabled extraction of generalizable latent features across subjects, highlighting corresponding anatomical patterns and tissue boundaries. As illustrated in \hyperlink{dmrif-figure5}{Fig. 5}, corresponding tissue structures showed visually consistent contrasts in the principal components of the features across the example subjects, e.g., ventricular regions appearing dark in component 1, deep white matter appearing dark in component 2, and the white/gray matter interface appearing bright in component 3. This consistency suggested that the features learned during pretraining captured stable structural information, providing a transferable basis for downstream computational dMRI tasks.

In subsequent experiments, we evaluated downstream task adaptation by fine-tuning task-specific decoders using the learned latent features for three distinct applications: (1) brain parcellation, which subdivides DWIs into anatomically segregated regions; (2) brain registration, which aligns DWI volumes across different subjects; and (3) cross-modal synthesis, which generates anatomical T1w images from DWIs. For each downstream task, from an input with six DWI volumes, the pre-trained dMRI-F encoder was employed to extract latent features, and a task-specific decoder was trained on a subset of 50 HCP-YA training subjects (see \hyperlink{dmrif-methods-downstream}{Methods}). In the following, we refer to these methods as dMRI-F-Parc, dMRI-F-Reg, and dMRI-F-Syn, respectively. Evaluation of each method was performed on both the high-quality HCP-YA data and the eight clinically typical test datasets, comparing their performance against representative SOTA methods in each task.

\subsubsection{Brain Parcellation into Anatomically Segregated Regions}

Brain parcellation is important in dMRI for region-specific quantification and anatomically accurate tractography. Parcellation directly from DWI bypasses the need for additional anatomical scans, avoids error-prone inter-modality registration, and accelerates processing time\cite{ref26,ref27}. Our proposed dMRI-F-Parc leverages learned feature representations to perform fine-scale parcellation of 101 anatomical regions corresponding to the Desikan-Killiany (DK) parcellation\cite{ref48} (see \hyperlink{dmrif-methods-parcellation}{Methods: Brain Parcellation}). We compared with two SOTA methods: (1) FastSurfer\cite{ref49}, which performs parcellation from T1w images and registers the parcellation to DWI; and (2) DDParcel\cite{ref27}, which uses dMRI-derived microstructure measures as inputs for parcellation (See \hyperlink{dmrif-methods-parcellation}{Methods} and Supplementary Table 4 for details).

As shown in \hyperlink{dmrif-figure6}{Fig. 6}, dMRI-F-Parc demonstrated consistent advantages across both the HCP-YA and clinically typical test datasets. In the HCP-YA data, it achieved the highest Dice scores relative to the ground-truth parcellation across cortical, subcortical, and white matter regions, significantly outperforming the compared methods (\hyperlink{dmrif-figure6}{Fig. 6a}). This improvement was most pronounced in cortical regions---notoriously difficult due to their complex cortical folding---where dMRI-F-Parc significantly outperformed the second-ranking DDParcel. Visual comparisons showed that dMRI-F-Parc parcellations more closely matched the ground-truth anatomical boundaries, particularly in the cortical regions (\hyperlink{dmrif-figure6}{Fig. 6b}). In the eight clinically acquired test datasets, dMRI-F-Parc showed robust generalization, achieving significantly lower relative standard deviation (RSD) of FA (a measure of regional parcellation homogeneity\cite{ref27}) than DDParcel in each dataset (\hyperlink{dmrif-figure6}{Fig. 6c}). Visual inspection supported these quantitative findings: dMRI-F-Parc produced regional tissue boundaries that were less displaced and better aligned with visible anatomical structures across heterogeneous datasets (\hyperlink{dmrif-figure6}{Fig. 6d}). Together, these results suggested that dMRI-F preserved parcellation-relevant anatomical information and enabled robust regional delineation directly from DWI data across various datasets.

\begin{figure}[!htbp]
\begin{center}
\hypertarget{dmrif-figure6}{}%
\includegraphics[width=\linewidth,height=0.62\textheight,keepaspectratio]{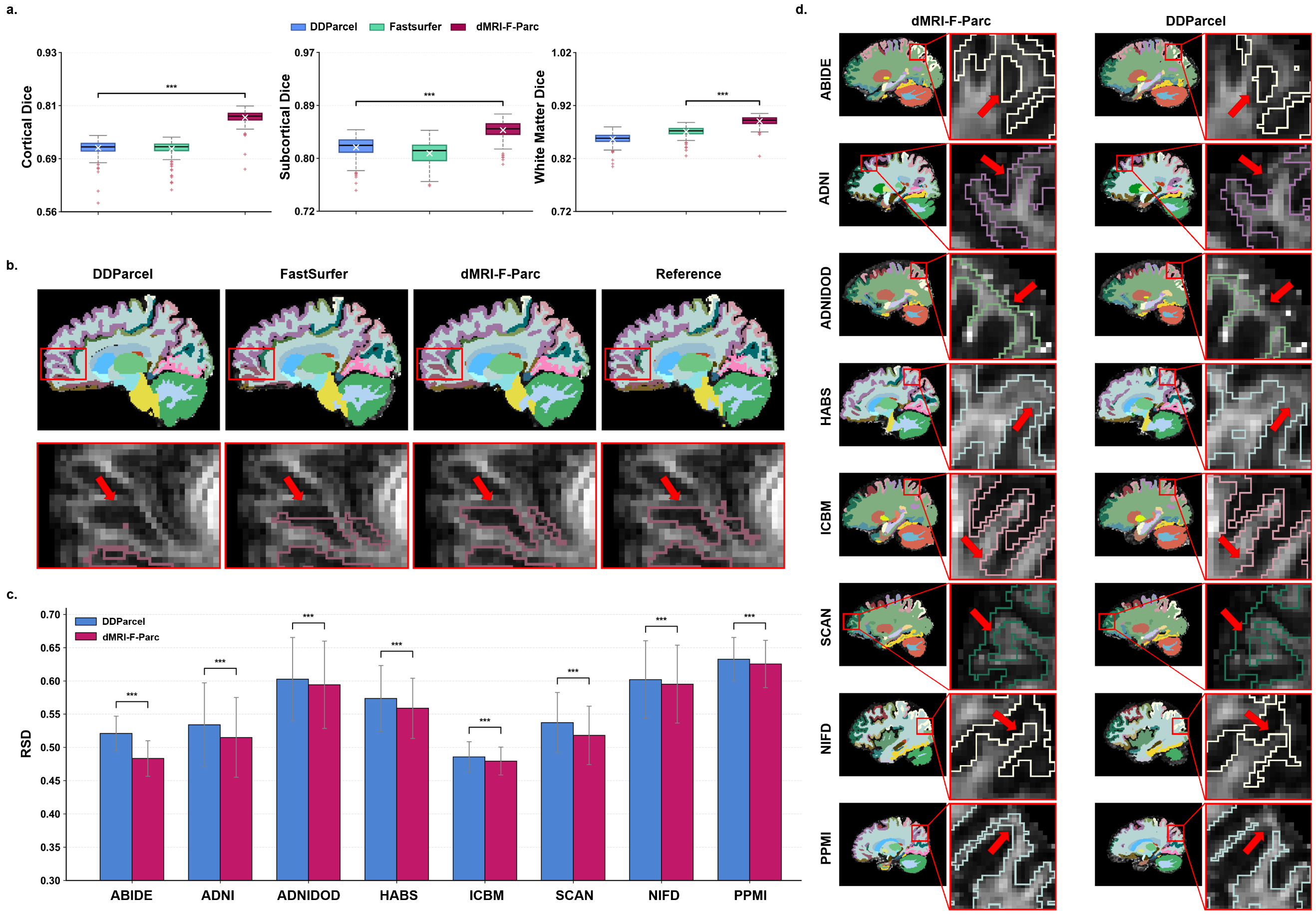}
\end{center}

\noindent\textbf{\textbf{Fig. 6 | dMRI-F-Parc enables accurate and generalizable brain parcellation directly from diffusion MRI.}}\par

\textbf{a,} Dice coefficients for cortical, subcortical, and white-matter parcellation against ground truth on the HCP-YA test set are compared across DDParcel, FastSurfer, and dMRI-F-Parc. dMRI-F-Parc achieved significantly higher Dice scores than DDParcel and FastSurfer across all region groups, with the largest gain in cortical regions. \textbf{ b, }Visualization of parcellation results from example subjects for each method, shown alongside the ground-truth reference, with enlarged views highlighting representative regions. dMRI-F-Parc more closely matched the tissue boundaries as appearing on the reference, particularly in cortical folding regions. Red arrows highlight representative regions where dMRI-F-Parc provides more accurate anatomical parcellation than the compared methods. \textbf{c, }Quantitative comparison on the eight clinical test datasets, where dMRI-F-Parc achieved significantly lower RSD than DDParcel across all datasets. \textbf{d,} Visual comparison of parcellation results on example subjects across datasets. The enlarged views show representative parcels, in which dMRI-F-Parc exhibits close visual alignment with the anatomical structures visible in the background FA image. Red arrows highlight representative regions where dMRI-F-Parc provides more accurate anatomical parcellation than DDParcel. \textit{Note:} For all box plots, the center line denotes the median, the box spans the interquartile range, and the whiskers extend to 1.5 times the interquartile range. White diagonal crosses denote the mean, whereas plus signs mark outliers beyond the whiskers. Bars show the cohort mean, and error bars indicate the standard deviation. Significance is shown for the comparisons marked by brackets: *\emph{p < 0.05}, **\emph{p < 0.01}, and ***\emph{p < 0.001}.
\end{figure}

\subsubsection{Brain Registration Between Subjects}

\begin{figure}[!htbp]
\begin{center}
\hypertarget{dmrif-figure7}{}%
\includegraphics[width=\linewidth,height=0.62\textheight,keepaspectratio]{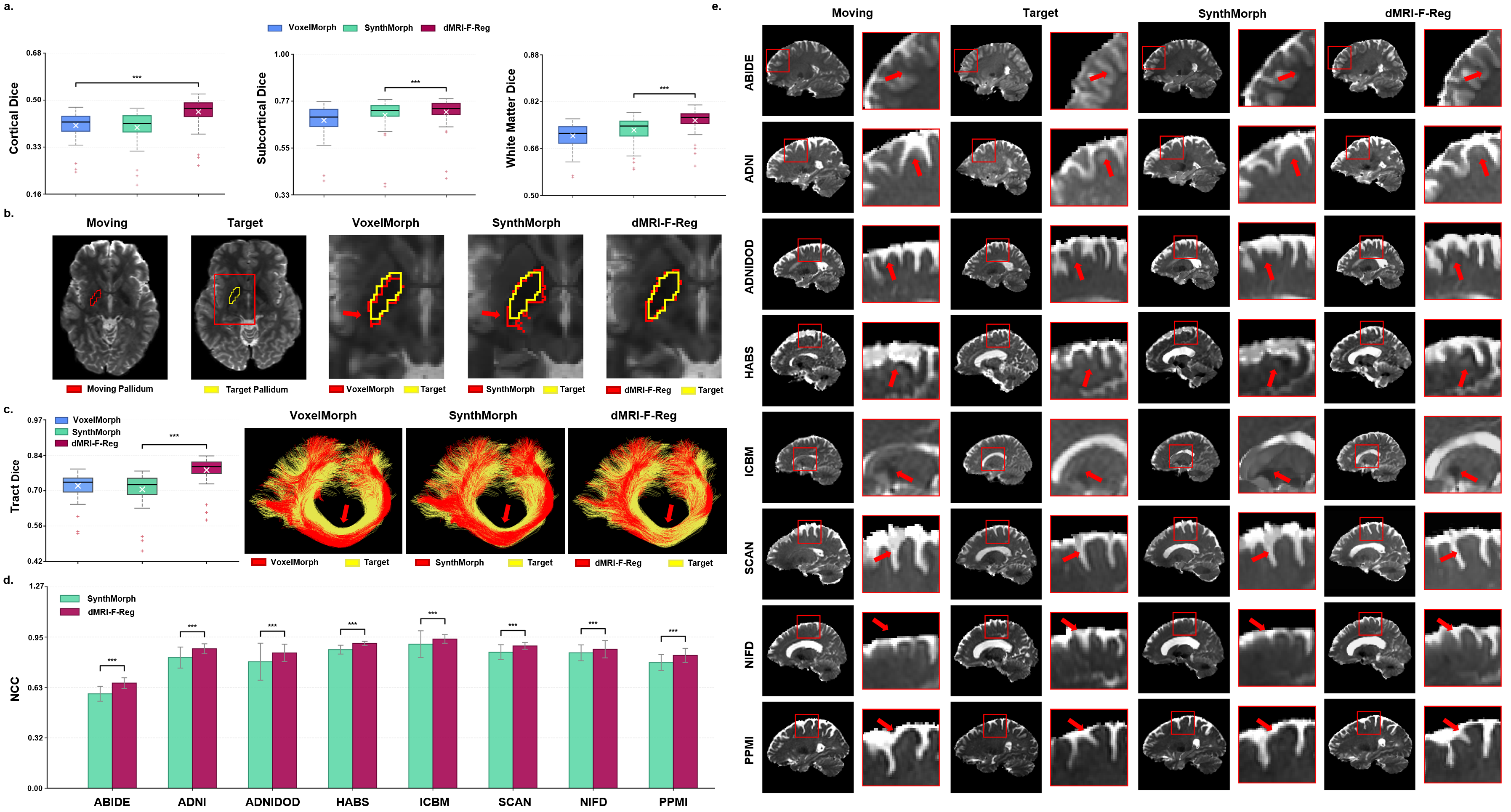}
\end{center}

\noindent\textbf{\textbf{F}\textbf{ig. 7 | dMRI-F}\textbf{-Reg improves inter-subject diffusion MRI registration across acquisition settings.}}\par

\textbf{a,} Performance on registering the corresponding anatomical structures on the HCP-YA test set. dMRI-F-Reg achieved significantly higher Dice scores than the compared methods across all region groups. \textbf{b,} Visualization of the registration results on an example moving and target HCP-YA subject pair. The enlarged views show the representative pallidum structure warped by each method (red) overlaid on the target (yellow), where dMRI-F-Reg yielded the closest spatial alignment. \textbf{c, }Performance on registering fiber tracts on the HCP-YA test set, where dMRI-F-Reg achieved the highest tract spatial Dice overlap. Visualization of an example corpus callosum (CC) tract after registration (in red) shows the closest spatial correspondence with the target tracts (in yellow). \textbf{d,} NCC between warped and fixed b0 images across the eight test datasets. dMRI-F-Reg achieved significantly higher NCC than SynthMorph in every dataset. \textbf{e,} Visualization of example moving, target, and warped b0 images. Enlarged views show improved local alignment with dMRI-F-Reg, and the arrows highlight representative regions where dMRI-F-Reg achieves closer anatomical alignment with the target images. \textit{Note:} For all box plots, the center line denotes the median, the box spans the interquartile range, and the whiskers extend to 1.5 times the interquartile range. White diagonal crosses denote the mean, whereas plus signs mark outliers beyond the whiskers. Bars show the cohort mean, and error bars indicate the standard deviation. Significance is shown for the comparisons marked by brackets: *\emph{p < 0.05}, **\emph{p < 0.01}, and ***\emph{p < 0.001}.
\end{figure}

Brain registration across subjects is a critical step in applications such as population-based analyses in health and disease and the construction of brain atlases. In DWI registration, the goal is to spatially align brain anatomical structures while preserving the consistency of local fiber orientations with the underlying white matter tract anatomy. Our proposed dMRI-F-Reg addresses this by training a registration model that takes learned features from subject pairs (moving and fixed) and predicts deformable fields to warp the moving DWI data into the fixed space (see \hyperlink{dmrif-methods-registration}{Methods: Brain Registration}). We compared with two SOTA methods: (1) VoxelMorph\cite{ref50}, which is a widely used unsupervised framework to learn deformation fields from input image pairs; and (2) SynthMorph\cite{ref51}, which is a synthesis-based registration method that generalizes robustly across contrasts, resolutions, and anatomies (See \hyperlink{dmrif-methods-registration}{Methods} and Supplementary Table 5 for details).

Quantitatively, dMRI-F-Reg achieved the highest Dice coefficient across cortical, subcortical, and white matter regions (\hyperlink{dmrif-figure7}{Fig. 7a}). Visual comparison on example HCP-YA subjects (\hyperlink{dmrif-figure7}{Fig. 7b}) shows that dMRI-F-Reg produced superior anatomical alignment, with warped anatomical regions more closely matching the fixed space than those from the compared method. The largest improvement was observed in the cortical regions, a particularly challenging area due to its highly folded geometry and substantial inter-subject variability. Furthermore, when warping fiber tracts (derived using TractSeg\cite{ref40}), dMRI-F-Reg produced the greatest spatial overlap (\hyperlink{dmrif-figure7}{Fig. 7c}), demonstrating more accurate alignment of tract-specific white-matter anatomy. Across all eight testing datasets, dMRI-F-Reg demonstrated statistically robust and spatially precise registration, achieving significantly higher Normalized Cross-Correlation (NCC) scores than SynthMorph (\hyperlink{dmrif-figure7}{Fig. 7d}). The qualitative improvements at tissue boundaries (\hyperlink{dmrif-figure7}{Fig. 7e}) indicate that dMRI-F-Reg better preserves structural similarity, likely due to its ability to model complex, non-linear deformations without over-smoothing---a common limitation in learning-based approaches like SynthMorph.

\subsubsection{Cross-Modal Synthesis of Anatomical T1w Images}

\begin{figure}[!htbp]
\begin{center}
\hypertarget{dmrif-figure8}{}%
\includegraphics[width=\linewidth,height=0.62\textheight,keepaspectratio]{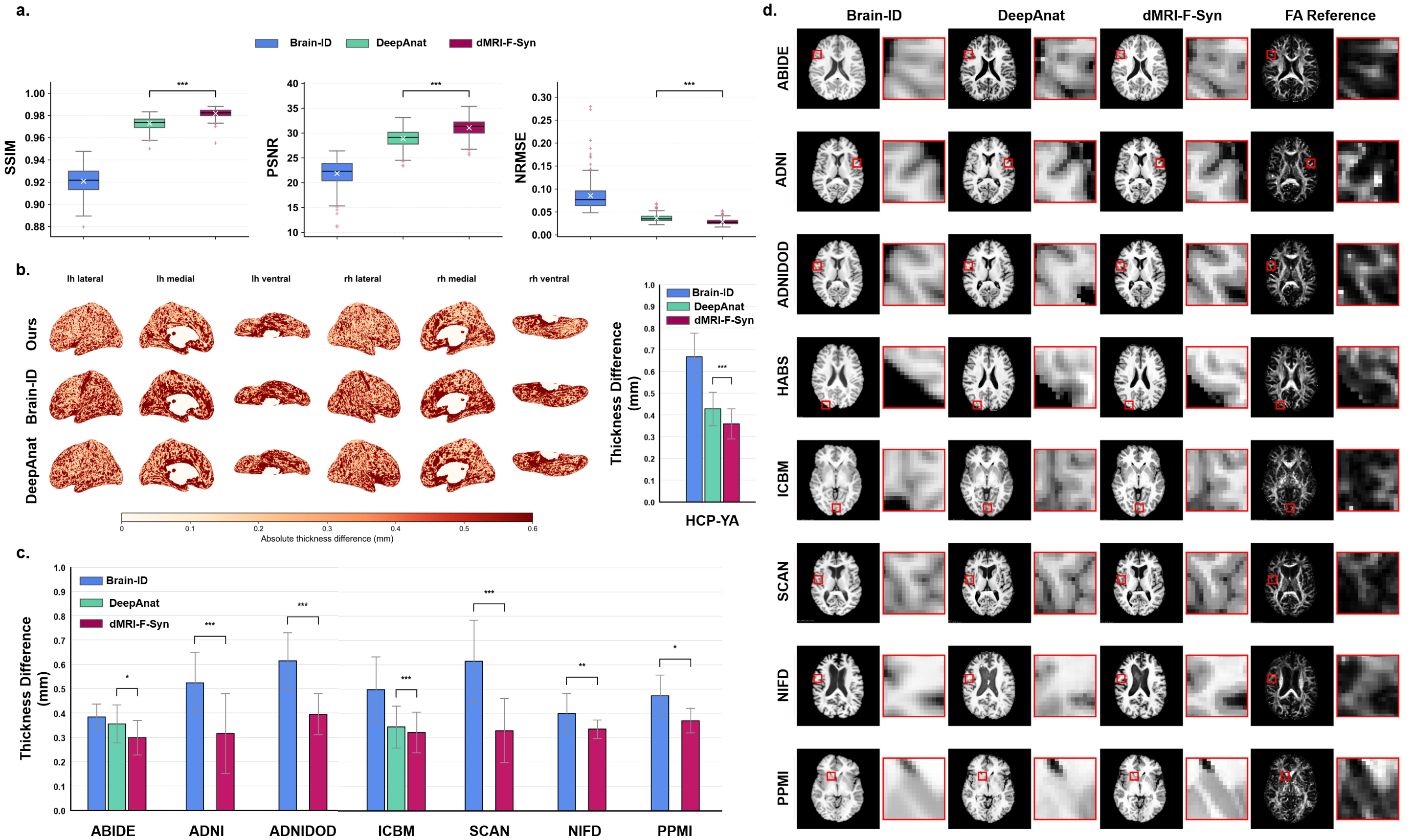}
\end{center}

\noindent\textbf{\textbf{Fig. 8 | }\textbf{dMRI-F-Syn e}\textbf{nables anatomically faithful synthesis of T1w images from diffusion MRI.}}\par

\textbf{a,} SSIM, PSNR, and NRMSE between synthesized and acquired T1w images in the HCP-YA test set. dMRI-F-Syn achieved significantly better results in each metric compared to Brain-ID and DeepAnat. \textbf{b,} Absolute cortical-thickness differences between surfaces reconstructed from the synthesized and the real T1w images in the HCP-YA data. Lateral, medial and ventral views of both hemispheres are shown. \textbf{c,} Mean absolute cortical-thickness differences between surfaces reconstructed from synthesized and acquired T1w images across the seven testing datasets. Bars represent the mean, and error bars represent the standard deviation. Missing DeepAnat results indicate datasets in which abnormal synthesized-image intensities prevented successful surface reconstruction. \textbf{d,} Representative T1w images synthesized by DeepAnat, Brain-ID, and dMRI-F-Syn across eight test datasets. Red boxes and enlarged views highlight local tissue boundaries and anatomical details. The corresponding reference FA maps are shown for anatomical localization. \textit{Note:} For all box plots, the center line denotes the median, the box spans the interquartile range, and the whiskers extend to 1.5 times the interquartile range. White diagonal crosses denote the mean, whereas plus signs mark outliers beyond the whiskers. Bars show the cohort mean, and error bars indicate the standard deviation. Significance is shown for the comparisons marked by brackets: *\emph{p < 0.05}, **\emph{p < 0.01}, and ***\emph{p < 0.001}.
\end{figure}

Cross-modal synthesis serves as an important tool for enriching dMRI with supplementary anatomical context, thereby supporting comprehensive neuroimaging analyses. While dMRI offers distinct information about water molecule diffusion properties, it lacks the explicit anatomical details as provided by T1w images. The proposed dMRI-F-Syn addresses this by learning mappings from the DWI-derived feature representations to the corresponding T1w data, enabling the prediction of synthetic T1w images from new DWI data (see \hyperlink{dmrif-methods-synthesis}{Methods: Cross-modal Synthesis}). We compared with two SOTA methods: (1) Brain-ID\cite{ref31}, which learns contrast-agnostic anatomical representations to synthesize T1w images for any imaging input (b0 is used in our experiment); (2) DeepAnat\cite{ref32}, which synthesizes T1w images directly from DWI using a U-Net and a hybrid generative adversarial network (see Supplementary Table 6 for details).

On the HCP-YA data, dMRI-F-Syn demonstrated superior T1w synthesis performance, yielding significantly higher SSIM, higher PSNR, and lower NRMSE against the ground-truth T1w images, than all compared methods (\hyperlink{dmrif-figure8}{Fig. 8a}). Moreover, the synthetic T1w images generated by dMRI-F-Syn enabled accurate cortical morphology estimation, as evidenced by significantly reduced cortical thickness differences relative to the ground-truth (\hyperlink{dmrif-figure8}{Fig. 8b}). Generalization across the seven test datasets was then assessed quantitatively by comparing cortical morphometry between the synthesized and acquired T1w images. dMRI-F-Syn achieved the lowest mean cortical-thickness difference in every dataset (\hyperlink{dmrif-figure8}{Fig. 8c}). (Note that DeepAnat failed to produce cortical surface reconstructions in several datasets due to abnormal intensity values, though it achieved relatively small cortical-thickness differences in the datasets where it succeeded.) Consistent with these quantitative findings, visual inspection showed that dMRI-F-Syn produced T1w images with sharper tissue boundaries and better-preserved local anatomical features (\hyperlink{dmrif-figure8}{Fig. 8d}). Collectively, these results demonstrate that dMRI-F-Syn effectively synthesizes T1w images that preserve anatomical information critical for both voxel-level reconstruction and downstream cortical-thickness estimation, even when applied to heterogeneous acquisitions with lower image quality.

\subsubsection{Outlook and Conclusion}

This work presents dMRI-F, which learns generalizable representations from sparse DWI inputs to support data enhancement and downstream analyses across heterogeneous acquisitions and populations. The enhanced DWIs could broaden the use of sparsely sampled datasets for studying structural connectivity and tissue microstructure across development, aging and disease. The ability to generate DWIs at arbitrary diffusion directions across multi shells could also support the training of other deep learning models with different sampling requirements. Beyond data enhancement, the learned representations could reduce reliance on additional anatomical scans and cross-modal registration in downstream analyses. These capabilities could enable more comprehensive diffusion imaging analyses from shorter acquisitions, potentially reducing patient burden and improving accessibility. Building on these capabilities, future work could extend dMRI-F to populations with tumours and other lesions, with particular attention to the preservation of pathological anatomy. Further studies could also explore the potential of the learned representations for predicting disease status, severity and treatment response, extending their use beyond the parcellation, registration and synthesis tasks evaluated here. Finally, integration with scanner reconstruction workflows and prospective evaluation using shortened acquisition protocols could help establish the reliability and practical value of dMRI-F in clinical settings.

\section{Methods}

The proposed dMRI-F is a foundation model for dMRI that performs DWI enhancement while supporting diverse downstream tasks across acquisition protocols and populations. It employs the popular self-supervised learning (SSL) framework with a Masked Autoencoder (MAE)\cite{ref52} design. SSL has emerged as a powerful paradigm for training foundation models by transforming unlabeled data into supervised learning signals through carefully designed pretext objectives\cite{ref53,ref54,ref55}. By exploiting the intrinsic structure of unlabeled data, SSL pre-training learns generalizable features that can be effectively transferred to a wide range of downstream tasks\cite{ref56,ref57}. Among various SSL strategies, masked reconstruction, as exemplified by MAE\cite{ref52}, has proven particularly effective. Its success relies on two closely related components: a pre-training task that defines what information should be hidden and recovered, and a model architecture capable of exploiting the remaining observations to infer the missing content. Further, the learned representations can be transferred to other computational tasks beyond pre-training. In the rest, we first describe the design of our pre-training task and the model architecture used to solve it, followed by the fine-tuning process for adapting the learned latent features to downstream tasks.

\subsection{Pretraining for DWI Enhancement}

\subsubsection{Pretraining Task}

The goal of our pre-training is to reconstruct missing DWI volumes from sparse input, thereby enabling effective DWI enhancement. In natural images and standard 3D medical images, reconstruction is commonly achieved by masking image patches or volumetric blocks. dMRI, however, presents a different setting: the acquired data form a 4D volume spanning three spatial dimensions and an additional diffusion-encoding dimension, where each DWI volume corresponds to a specific q-space sampling point determined by the gradient direction (b-vector) and strength (b-value) (see \hyperlink{dmrif-figure1}{Fig. 1a}). Following the masked reconstruction principle, we formulate the pre-training task as angular masked reconstruction, which recovers unobserved diffusion signals from sparse observations.

Specifically, let $X\in\mathbb{R}^{H\times W\times D\times G}$ denote the complete set of DWI signals acquired at non-zero b-values, where $H$, $W$ and $D$ are the spatial dimensions, and $G$ is the diffusion-encoding dimension (i.e., the total number of acquired DWI volumes across all gradient directions and b-value shells). The b0 volumes are excluded from both the pretraining inputs and reconstruction targets but are retained for subsequent analyses, including microstructural estimation, FOD estimation, and tractography. We define a masking operator $M$ that selects a sparse subset of q-space coordinates $q_{s}$, yielding the q-space subsampled input volumes $\overline{X}=M(X)$. The pre-training task is then defined as:

\[\hat{X}(q_{t})=F_{\theta }(q_{s}, \overline{X}, q_{t}),\]
\noindent where $q_{t}$ denotes the target q-space coordinate, and $F_{\theta}$ learns to predict the missing DWIs from the sparse input. In our pre-training setting, the input $\overline{X}$ contains only 6 arbitrarily selected DWI volumes at b = 1,000\,s/mm\textsuperscript{2}, while the reconstruction target comprises the full DWI data, including 270 DWIs across b = 1,000, 2,000, and 3,000\,s/mm\textsuperscript{2}.

\subsubsection{Network Architecture}

To solve the pre-training task, we design an SSL network based on the MAE architecture to recover the masked DWIs. In conventional spatial masked reconstruction tasks, the objective is typically to recover masked patches from the observed context by exploiting spatial dependencies within the image. Our angular masked reconstruction task, however, requires modeling additional context, because DWIs acquired under different q-space locations are not independent but rather reflect the same underlying tissue anatomy. Recovering the masked DWIs thus requires jointly modeling the relationships among spatial coordinates, underlying tissue anatomy, and q-space coordinates, motivating an architecture that integrates anatomical context with spatial and q-space coordinates.

Implicit Neural Representations (INRs)\cite{ref58} are particularly suited to this requirement. INRs are a class of neural networks that learn continuous, coordinate-based mappings from input coordinates to output signals, enabling flexible and high-fidelity representation of complex data. Neural radiance fields (NeRF)\cite{ref59} is a landmark implementation of the INR paradigm, where an INR is trained to represent a 3D scene as a continuous function of spatial coordinates and viewing directions, enabling scene prediction from novel viewpoints. The design of NeRF inspires our approach to modeling DWI signals as continuous functions of spatial and q-space coordinates. In related work, recent dMRI studies have demonstrated the success of INRs for continuously modeling fiber orientation distributions and data super-resolution\cite{ref60,ref61,ref62,ref63}. However, these approaches typically require subject-specific optimization and are not designed for direct transfer across acquisition protocols or heterogeneous datasets. In our study, we developed dMRI-F as a unified pretrained framework that takes sparsely sampled DWI volumes as inputs and reconstructs complete DWI volumes at unobserved q-space locations. Through q-space masked reconstruction over a large cohort of training subjects, its encoder learns to extract structural representations consistently from unseen subjects and heterogeneous datasets without subject-specific optimization.

Specifically, we adopt an encoder--decoder architecture that learns a coordinate-conditioned mapping across subjects. Given the input DWI data $\overline{X}$ consisting of $K$ volumes, the encoder $E_{\phi }$ extracts a structural feature field $\Phi$ that captures anatomical information shared across DWIs. At each spatial location $x$, the implicit mapping network within the decoder models the relationship among the local structural feature $\Phi (x)$, the $K$ corresponding q-space coordinates ${\eta (q_{s})}_{s=1}^{K}$, and the target q-space coordinate $q_{t}$. This formulation allows the model to adapt its prediction to subject-specific anatomy through the encoded feature field while avoiding per-subject optimization. The implicit mapping module $G_{\psi }$ then predicts a voxel-wise latent code for each target spatial and q-space coordinate:

\[z(x, q_{t})=G_{\psi }(\gamma (x), \eta (q_{t}); \Phi (x),{\eta (q_{s})}_{s=1}^{K}),\]
\noindent where $\gamma (\cdot )$ and $\eta (\cdot )$ denote the spatial and angular coordinate embeddings, respectively.

For the angular coordinate, we encode the queried diffusion direction using a real spherical-harmonics (SH) expansion:

\[\eta (q)=\mathrm{SH}_{L}(q)=[Y_{0,0}^{\mathrm{real}}(q),Y_{1,-1}^{\mathrm{real}}(q), ..., Y_{l,m}^{\mathrm{real}}(q),Y_{L-1,L-1}^{\mathrm{real}}(q)],\]
\noindent where $Y_{l,m}^{\mathrm{real}}(q)$ denotes the real SH basis function. We set $L = 4$, which provides a compact and smooth representation of angular variation on the sphere. For the spatial coordinate, we use the global voxel coordinate $x$ in the DWI volume, normalize each dimension to $[-1, 1]$, and apply a sinusoidal positional encoding:

\[\gamma (x)=\operatorname{SIN}(x)=[x,\sin(2^{0}\pi x),\cos(2^{0}\pi x),...,\sin(2^{N-1}\pi x),\cos(2^{N-1}\pi x)],\]
\noindent where $N$ denotes the number of spatial frequency bands.

The resulting latent codes are assembled and refined by a volumetric decoder to reconstruct the high-resolution DWI at the target q-space location.

\subsubsection{Pretraining Loss}

To preserve fine-grained image details, the model is optimized using a combination of a reconstruction loss and a perceptual loss:

\[L=L_{1}^{\mathrm{DWI}}+\lambda L_{\mathrm{perc}}^{\mathrm{DWI}},\]
\noindent where $\lambda$ denotes the weighting factor for the perceptual loss and is set to 0.01. The voxel-wise reconstruction loss is computed as:

\[L_{1}^{\mathrm{DWI}}=\lVert\hat{X}_{q_{t}}-X_{q_{t}}\rVert_{1},\]
\noindent where $\hat{X}_{q_{t}}$ and $X_{q_{t}}$ denote the reconstructed and reference DWIs at the target q-space coordinate $q_{t}$, respectively. The perceptual loss is defined as

\[L_{\mathrm{perc}}^{\mathrm{DWI}}=\sum_{l\in S}\lVert\sigma _{l}(\hat{X}_{q_{t}})-\sigma _{l}(X_{q_{t}})\rVert_{1},\]
\noindent where $\sigma _{l}(\cdot )$ denotes the feature representation extracted at layer $l$, and $S$ is the set of selected feature layers.

\subsection{Fine-Tuning for Downstream Tasks}
\hypertarget{dmrif-methods-downstream}{}%

Following q-space masked reconstruction pre-training, we adapted the learned feature representations to three downstream tasks: brain parcellation, brain registration, and cross-modal synthesis. For each task, the pre-trained encoder served as a feature extractor. At each training iteration, six DWI volumes were randomly selected from the b = 1,000\,s/mm\textsuperscript{2} shell of each subject and used as encoder inputs. A task-specific decoder was then fine-tuned on a subset of 50 HCP-YA subjects randomly selected from the pre-training set. For the decoder, we adopted a U-Net architecture optimized with a task-specific loss function. To evaluate performance, we compared against SOTA methods in each task, on the high-quality HCP-YA dataset and the eight independently acquired test datasets with clinically typical acquisitions.

\subsubsection{Brain Parcellation}
\hypertarget{dmrif-methods-parcellation}{}%

For brain parcellation, the goal was to predict a voxel-wise anatomical label from the input DWI volumes. In our study, we trained a parcellation decoder that mapped the extracted features to a parcellation comprising 101 anatomical labels corresponding to the FreeSurfer Desikan--Killiany (DK) parcellation\cite{ref48}. The decoder was trained using a combination of cross-entropy and Dice losses:

\[L_{\mathrm{Parc}}=L_{\mathrm{CE}}+L_{\mathrm{Dice}},\]
\noindent where the cross-entropy term $L_{\mathrm{CE}}$ provided voxel-wise supervision for anatomical classification, and the Dice term $L_{\mathrm{Dice}}$ promoted spatial overlap between the predicted and reference regions while reducing the influence of class imbalance.

We compared the SOTA brain parcellation methods, FastSurfer\cite{ref49} and DDParcel\cite{ref27}. FastSurfer aggregates information from neighboring T1w slices and combines predictions across three anatomical planes to generate whole-brain parcellations. DDParcel integrates complementary microstructural information from diffusion-derived parameter maps at multiple levels of the network to generate high-quality anatomical parcellations. For evaluation, we adopted strategies from a previous DWI parcellation study\cite{ref27}. For HCP-YA, the advanced acquisition protocol supported accurate registration between anatomical T1w images and DWIs. We therefore used FreeSurfer parcellations derived from T1w images and registered to DWI space as the reference, and quantified performance using Dice. For the testing datasets with clinically typical acquisitions, lower image quality and less reliable cross-modality registration prevented the construction of comparably accurate reference parcellations in DWI space. We therefore assessed regional homogeneity using the relative standard deviation (RSD) of FA, defined as the within-region standard deviation divided by its mean. Lower RSD values indicate greater anatomical and microstructural coherence within the predicted regions.

\subsubsection{Brain Registration}
\hypertarget{dmrif-methods-registration}{}%

For brain registration, the goal was to estimate a non-linear deformation field aligning the DWI data of a moving subject to a fixed subject. To estimate this field, the features extracted from the moving and fixed subjects were jointly processed by a registration network. The predicted deformation field was then applied to the entire moving-subject DWI series to warp it into fixed-subject space. The network was optimized using image-similarity and deformation-smoothness losses:

\[L_{\mathrm{Reg}}=L_{\mathrm{MSE}}+L_{\mathrm{grad}},\]
\noindent where the MSE term $L_{\mathrm{MSE}}$ promotes intensity agreement between the warped moving and fixed b0 images, and the gradient term $L_{\mathrm{grad}}$ constrains spatial variations in the displacement field to encourage smooth and anatomically plausible deformations.

Registration performance was evaluated against the SOTA registration methods: VoxelMorph\cite{ref50} and SynthMorph\cite{ref51}. VoxelMorph learns a direct mapping from image pairs to dense deformation fields using a convolutional neural network, enabling rapid deformable registration in a single forward pass. SynthMorph learns registration from images synthesized with diverse contrasts and deformations, using a label-overlap objective to promote anatomical alignment independent of image intensities. This training strategy supports registration across MRI contrasts, including those not encountered during training. Registration performance was evaluated at complementary anatomical and white-matter tract levels. For HCP-YA, FreeSurfer-derived cortical, subcortical, and white-matter parcellations from each moving subject were warped into fixed-subject space using the deformation field predicted by each method. Their spatial overlap with the corresponding fixed-subject parcellations was quantified using the Dice coefficient. To assess tract alignment across the complete diffusion acquisition, each deformation field was also applied to all moving-subject FOD volumes. TractSeg was then applied to the warped moving-subject FODs and the fixed-subject FODs, and the resulting tract segmentations were compared using the same metrics. For the eight test datasets, registration performance was assessed using normalized cross-correlation (NCC) between the warped moving and fixed b0 images, complemented by visual inspection of local anatomical correspondence.

\subsubsection{Cross-Modal Synthesis}
\hypertarget{dmrif-methods-synthesis}{}%

In cross-modal synthesis, anatomical T1-weighted (T1w) images were predicted directly from DWI volumes. The synthesis decoder mapped the extracted features to the corresponding spatially aligned T1w image and was optimized using a combination of voxel-wise and perceptual losses:

\[L_{\mathrm{Syn}}=L_{1}^{\mathrm{T1w}}+\lambda L_{\mathrm{perc}}^{\mathrm{T1w}},\]
\noindent where $\lambda$ is set to 0.01. The $L_{1}^{\mathrm{T1w}}$ term enforces voxel-wise intensity agreement between the synthesized and reference T1-weighted images, whereas the perceptual term promotes the preservation of higher-level anatomical features.

For T1w image synthesis, we compared with the Brain-ID\cite{ref31} and DeepAnat\cite{ref32}. Brain-ID learns contrast-agnostic anatomical representations from synthetic inputs with varying contrasts, resolutions, and corruption levels, supporting T1w synthesis through a lightweight task-specific output layer. In our comparison, b0 images were used as the input. DeepAnat synthesizes T1w images directly from diffusion data using generative adversarial network architectures. The synthesized images retain spatial correspondence with the diffusion data, supporting anatomical segmentation. For HCP-YA, accurately aligned T1w reference images enabled voxel-wise evaluation of the three methods using SSIM, PSNR and NRMSE. Preservation of cortical morphology was further assessed by applying FastSurfer to the synthesized T1w images and comparing the resulting cortical thickness maps with the FreeSurfer-derived HCP-YA references. Cortical thickness differences were also visualized for a representative subject. For the independent datasets, inconsistent alignment between the acquired T1w images and DWIs precluded reliable voxel-wise comparison. We therefore applied FastSurfer to both the synthesized and acquired T1w images and compared the resulting cortical thickness estimates to assess anatomical fidelity at the cortical-surface level.

\subsection{Datasets and Implementation}

\subsubsection{Training Data}

We selected the HCP-YA data as the pre-training dataset and included 200 young healthy adults (age, 28.84\,\ensuremath{\pm}\,3.51 years; 118 females and 82 males). Of these, 180 subjects were used for training and 20 for validation (see \hyperlink{dmrif-extended-figure1}{Extended Data Fig. 1} for the ablation study on training data size). Imaging was performed on a customized Siemens 3\,T Connectome Skyra scanner using the high-quality HCP acquisition protocol. The dMRI data were acquired with TE\,=\,89.5\,ms, TR\,=\,5,520\,ms and a 1.25-mm isotropic voxel size. Each subject had 288 volumes, comprising 18 b0 images and 270 DWIs distributed equally across three shells at b = 1,000, 2,000, and 3,000\,s/mm\textsuperscript{2}. Images were acquired with both left-to-right and right-to-left phase encoding to facilitate correction of echo-planar imaging distortions. The b = 1,000\,s/mm\textsuperscript{2} shell was used to construct the sparse input, reflecting its broad availability in conventional and legacy dMRI protocols, whereas the complete multi-shell acquisition served as the prediction target. The corresponding T1w images were acquired with TE\,=\,2.14\,ms, TR\,=\,2,400\,ms and a 0.7-mm isotropic voxel size. We used data processed with the HCP minimal preprocessing pipelines, which corrected EPI and eddy-current distortions, subject motion and gradient nonlinearities, and registered the dMRI data to the native T1w structural space at 1.25-mm isotropic resolution. This processing provided precise within-subject alignment between the dMRI and T1-weighted images. The minimally preprocessed DWIs were further denoised using Patch2Self\cite{ref24} to suppress residual noise and improve the clarity of anatomical structures.

\subsubsection{Testing Data}
\hypertarget{dmrif-methods-testing-data}{}%

To evaluate the generalizability of dMRI-F beyond the pre-training data, we used the HCP-YA test set together with eight test datasets: ABIDE\cite{ref64}, ADNI\cite{ref46}, ADNI-DOD\cite{ref65}, HABS\cite{ref66}, ICBM\cite{ref67}, SCAN\cite{refSCAN}, NIFD\cite{ref68}, and PPMI\cite{ref69}. Together, the HCP-YA test set and the eight test datasets span a broad age range and include healthy participants as well as individuals with neurodevelopmental, neurodegenerative, psychiatric, and trauma-related conditions, with data acquired using diverse scanners and diffusion protocols. Detailed sample sizes, demographic and diagnostic characteristics, scanner information, and acquisition parameters are provided in Supplementary Table 3. The original studies were approved by their respective institutional review boards or ethics committees, and all de-identified data were used in accordance with the corresponding data-use agreements. Participants in the HCP-YA test set and the eight test datasets did not overlap with those used for pre-training. Data from the HCP-YA test set were processed using the same HCP minimal preprocessing pipelines\cite{ref70} described above. For the test datasets available in DICOM format, the images were converted to NIfTI while preserving the associated b-values and diffusion gradient directions. All dMRI data from the eight test datasets underwent correction of motion- and eddy-current-induced distortions using DTIPrep (v1.2.11)\cite{ref71}. Brain masks were generated using CNN-Diffusion-MRIBrain-Segmentation (v0.3)\cite{ref72}. Patch2Self was then applied to the complete diffusion series to generate the denoised reference data, whereas the six sparse input DWIs were selected from the corresponding undenoised acquisitions. Finally, all dMRI data from the eight test datasets underwent visual quality assessment.

\subsection{Implementation}

The input DWIs were min--max normalized using the intensities across DWIs for each subject. During pretraining, we randomly reduced the spatial resolution of the input DWIs as data augmentation to promote the learning of more generalizable representations. All models except the registration model received patches of 64 \ensuremath{\times} 64 \ensuremath{\times} 64 voxels as input. For registration, the extracted feature patches were first reassembled into whole-brain feature volumes to preserve spatial continuity, then processed by the registration network. The structural encoder, volumetric decoder, and task-specific decoders for all downstream tasks were implemented using 3D U-Net architectures. The learning rate was set to $1\times10^{-4}$. All models were implemented in PyTorch, and training and inference were performed on NVIDIA RTX 3090 GPUs.

\section*{Data Availability}

The data used in this study are available from the Human Connectome Project Young Adult (HCP-YA; https://www.humanconnectome.org/study/hcp-young-adult/data-releases), Autism Brain Imaging Data Exchange (ABIDE; https://fcon\_1000.projects.nitrc.org/indi/abide/), Alzheimer's Disease Neuroimaging Initiative (ADNI; https://adni.loni.usc.edu/data-samples/adni-data/), Department of Defense Alzheimer's Disease Neuroimaging Initiative (DOD ADNI; https://ida.loni.usc.edu/collaboration/access/appApply.jsp?project=ADNIDOD), Harvard Aging Brain Study (HABS; https://www.synapse.org/habs), International Consortium for Brain Mapping (ICBM; https://ida.loni.usc.edu/collaboration/access/appApply.jsp?project=ICBM), Standardized Centralized Alzheimer's and Related Dementias Neuroimaging initiative (SCAN; https://scan.naccdata.org), Neuroimaging in Frontotemporal Dementia study (NIFD; https://ida.loni.usc.edu/collaboration/access/appApply.jsp?project=NIFD), and Parkinson's Progression Markers Initiative (PPMI; https://www.ppmi-info.org/access-data-specimens/download-data/). Access to controlled datasets is subject to registration, approval, and the respective data-use agreements. No new datasets were generated in this study.

\section*{Code Availability}

The code for pretraining, fine-tuning, and evaluation of the foundation model is available in GitHub at https://github.com/XiZhu-UE/dMRI-F.


\section*{Acknowledgment}

The NACC database is funded by NIA/NIH Grant U24 AG072122. SCAN is a multi-institutional project that was funded as a U24 grant (AG067418) by the National Institute on Aging in May 2020. Data collected by SCAN and shared by NACC are contributed by the NIA-funded ADRCs as follows:

Arizona Alzheimer's Center - P30 AG072980 (PI: Eric Reiman, MD); R01 AG069453 (PI: Eric Reiman (contact), MD); P30 AG019610 (PI: Eric Reiman, MD); and the State of Arizona which provided additional funding supporting our center; Boston University - P30 AG013846 (PI Neil Kowall MD); Cleveland ADRC - P30 AG062428 (James Leverenz, MD); Cleveland Clinic, Las Vegas - P20AG068053; Columbia - P50 AG008702 (PI Scott Small MD); Duke/UNC ADRC - P30 AG072958; Emory University - P30AG066511 (PI Levey Allan, MD, PhD); Indiana University - R01 AG19771 (PI Andrew Saykin, PsyD); P30 AG10133 (PI Andrew Saykin, PsyD); P30 AG072976 (PI Andrew Saykin, PsyD); R01 AG061788 (PI Shannon Risacher, PhD); R01 AG053993 (PI Yu-Chien Wu, MD, PhD); U01 AG057195 (PI Liana Apostolova, MD); U19 AG063911 (PI Bradley Boeve, MD); and the Indiana University Department of Radiology and Imaging Sciences; Johns Hopkins - P30 AG066507 (PI Marilyn Albert, Phd.); Mayo Clinic - P50 AG016574 (PI Ronald Petersen MD PhD); Mount Sinai - P30 AG066514 (PI Mary Sano, PhD); R01 AG054110 (PI Trey Hedden, PhD); R01 AG053509 (PI Trey Hedden, PhD); New York University - P30AG066512-01S2 (PI Thomas Wisniewski, MD); R01AG056031 (PI Ricardo Osorio, MD); R01AG056531 (PIs Ricardo Osorio, MD; Girardin Jean-Louis, PhD); Northwestern University - P30 AG013854 (PI Robert Vassar PhD); R01 AG045571 (PI Emily Rogalski, PhD); R56 AG045571, (PI Emily Rogalski, PhD); R01 AG067781, (PI Emily Rogalski, PhD); U19 AG073153, (PI Emily Rogalski, PhD); R01 DC008552, (M.-Marsel Mesulam, MD); R01 AG077444, (PIs M.-Marsel Mesulam, MD, Emily Rogalski, PhD); R01 NS075075 (PI Emily Rogalski, PhD); R01 AG056258 (PI Emily Rogalski, PhD); Oregon Health and Science University - P30 AG066518 (PIs Lisa Silbert, MD, Kevin Duff, PhD); R56 AG074321 (PI Jeffrey Kaye, MD); Rush University - P30 AG010161 (PI David Bennett MD); Stanford - P30AG066515; P50 AG047366 (PI Victor Henderson MD MS); University of Alabama, Birmingham - P20; University of California, Davis - P30 AG10129 (PI Charles DeCarli, MD); P30 AG072972 (PI Charles DeCarli, MD); University of California, Irvine - P50 AG016573 (PI Frank LaFerla PhD); University of California, San Diego - P30AG062429 (PI James Brewer, MD, PhD); University of California, San Francisco - P30 AG062422 (Rabinovici, Gil D., MD); University of Kansas - P30 AG035982 (Russell Swerdlow, MD); University of Kentucky - P30 AG028283-15S1 (PIs Linda Van Eldik, PhD and Brian Gold, PhD); University of Michigan ADRC - P30AG053760 (PI Henry Paulson, MD, PhD) P30AG072931 (PI Henry Paulson, MD, PhD) Cure Alzheimer's Fund 200775 - (PI Henry Paulson, MD, PhD) U19 NS120384 (PI Charles DeCarli, MD, University of Michigan Site PI Henry Paulson, MD, PhD) R01 AG068338 (MPI Bruno Giordani, PhD, Carol Persad, PhD, Yi Murphey, PhD) S10OD026738-01 (PI Douglas Noll, PhD) R01 AG058724 (PI Benjamin Hampstead, PhD) R35 AG072262 (PI Benjamin Hampstead, PhD) W81XWH2110743 (PI Benjamin Hampstead, PhD) R01 AG073235 (PI Nancy Chiaravalloti, University of Michigan Site PI Benjamin Hampstead, PhD) 1I01RX001534 (PI Benjamin Hampstead, PhD) IRX001381 (PI Benjamin Hampstead, PhD); University of New Mexico - P20 AG068077 (Gary Rosenberg, MD); University of Pennsylvania - State of PA project 2019NF4100087335 (PI David Wolk, MD); Rooney Family Research Fund (PI David Wolk, MD); R01 AG055005 (PI David Wolk, MD); University of Pittsburgh - P50 AG005133 (PI Oscar Lopez MD); University of Southern California - P50 AG005142 (PI Helena Chui MD); University of Washington - P50 AG005136 (PI Thomas Grabowski MD); University of Wisconsin - P50 AG033514 (PI Sanjay Asthana MD FRCP); Vanderbilt University - P20 AG068082; Wake Forest - P30AG072947 (PI Suzanne Craft, PhD); Washington University, St. Louis - P01 AG03991 (PI John Morris MD); P01 AG026276 (PI John Morris MD); P20 MH071616 (PI Dan Marcus); P30 AG066444 (PI John Morris MD); P30 NS098577 (PI Dan Marcus); R01 AG021910 (PI Randy Buckner); R01 AG043434 (PI Catherine Roe); R01 EB009352 (PI Dan Marcus); UL1 TR000448 (PI Brad Evanoff); U24 RR021382 (PI Bruce Rosen); Avid Radiopharmaceuticals / Eli Lilly; Yale - P50 AG047270 (PI Stephen Strittmatter MD PhD); R01AG052560 (MPI: Christopher van Dyck, MD; Richard Carson, PhD); R01AG062276 (PI: Christopher van Dyck, MD); Florida - P30AG066506-03 (PI Glenn Smith, PhD); P50 AG047266 (PI Todd Golde MD PhD)

\section*{Funding}

This work is in part supported by the National Natural Science Foundation of China (No. 62371107) and the National Key R\&D Program of China (No. 2023YFE0118600).

\section*{Author Contribution}

Xi Zhu contributed to conceptualization, data curation, investigation, methodology, supervision, validation, visualization, and writing of the original draft, and reviewed and edited the manuscript. Yijie Li contributed to data curation and reviewed and edited the manuscript. Xiaofan Wang, Wei Zhang, Huafu Chen and Ce Zhu reviewed and edited the manuscript. Fan Zhang contributed to methodology, project administration, supervision, and writing of the original draft, and reviewed and edited the manuscript.

\section*{Competing Interests}

The authors declare no competing interests.

\section{Extended Data}

\subsection{Extended Data Table 1 | Ablation study of spherical harmonics encoding and the volumetric decoder.}

\begin{center}
\hypertarget{dmrif-extended-table1}{}%
\resizebox{0.78\textwidth}{!}{%
\begin{tabular}{lcccc}
\toprule
SH-Encoding & Decoder & SSIM\ensuremath{\uparrow} & PSNR\ensuremath{\uparrow} & NRMSE\ensuremath{\downarrow} \\
\midrule
\ensuremath{\surd} & - & 0.9768\ensuremath{\pm}0.0056 & 35.8345\ensuremath{\pm}1.651 & 0.0164\ensuremath{\pm}0.0031 \\
- & \ensuremath{\surd} & 0.9721\ensuremath{\pm}0.0071 & 34.8715\ensuremath{\pm}1.344 & 0.0183\ensuremath{\pm}0.0028 \\
\ensuremath{\surd} & \ensuremath{\surd} & \textbf{0.9827\ensuremath{\pm}0.0020} & \textbf{37.1959\ensuremath{\pm}0.410} & \textbf{0.01387\ensuremath{\pm}0.006} \\
\bottomrule
\end{tabular}%
}
\end{center}

DWI prediction performance on the HCP-YA test set was evaluated after separately removing the SH encoding or the volumetric decoder. In the variant without SH encoding, q-space coordinates were passed directly to the decoder without SH transformation. In the variant without the volumetric decoder, the decoder's implicit mapping network directly predicted voxel-wise signal intensities to form the target DWI, rather than generating latent codes for subsequent volumetric decoding. The complete model achieved higher SSIM and PSNR and lower NRMSE than either ablated variant, supporting the contribution of both components to DWI prediction accuracy.

\clearpage

\begin{figure}[!htbp]
\begin{center}
\hypertarget{dmrif-extended-figure1}{}%
\includegraphics[width=\linewidth,height=0.62\textheight,keepaspectratio]{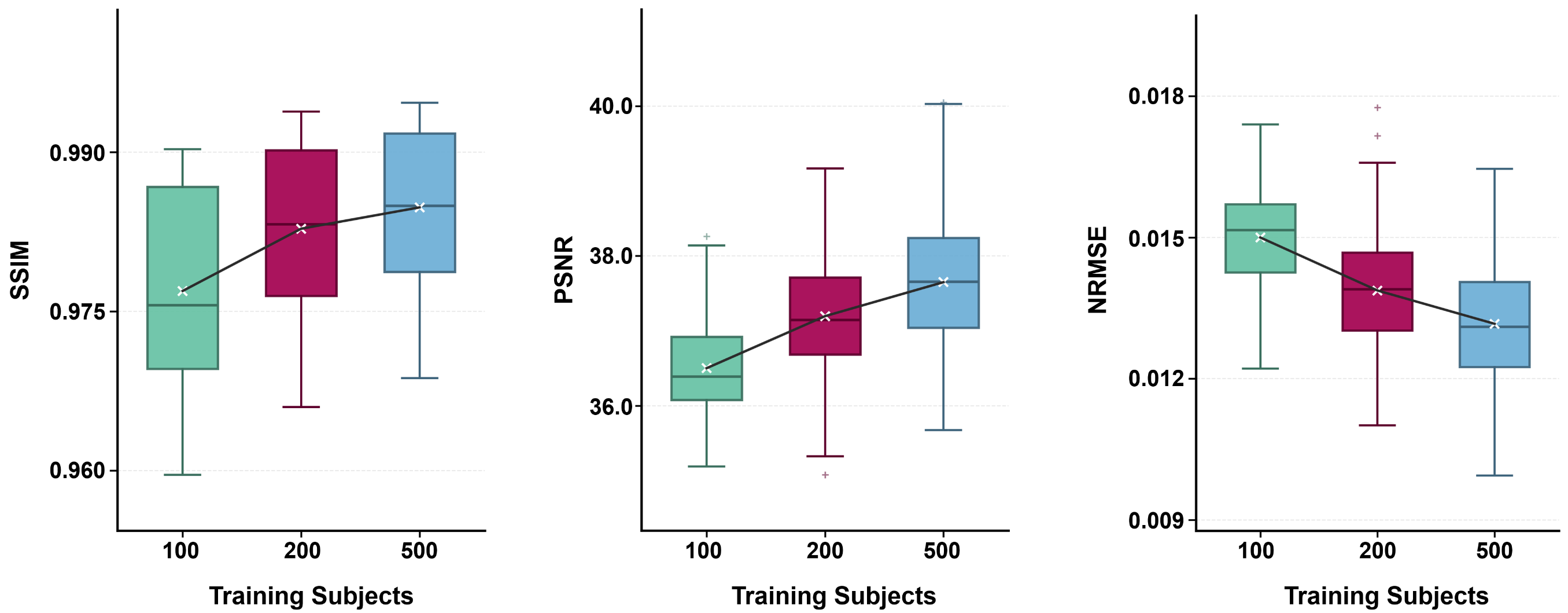}
\end{center}

\noindent\textbf{\textbf{Extended Data Fig. 1 | Effect of training data size on DWI enhancement performance. }}\par

To examine how the amount of training data affects DWI enhancement, we varied the training data size from 100 to 200 and 500 subjects and evaluated the resulting dMRI-F models on the HCP-YA test set. Performance improved with larger training datasets, as reflected by higher SSIM and PSNR and lower NRMSE. For all box plots, the center line denotes the median, the box spans the interquartile range, and the whiskers extend to the most extreme values within 1.5 times the interquartile range of the box. White crosses indicate means, connected by black lines to illustrate trends across training cohort sizes; plus signs denote outliers beyond the whiskers.
\end{figure}

\begin{figure}[!htbp]
\begin{center}
\hypertarget{dmrif-extended-figure2}{}%
\includegraphics[width=\linewidth,height=0.62\textheight,keepaspectratio]{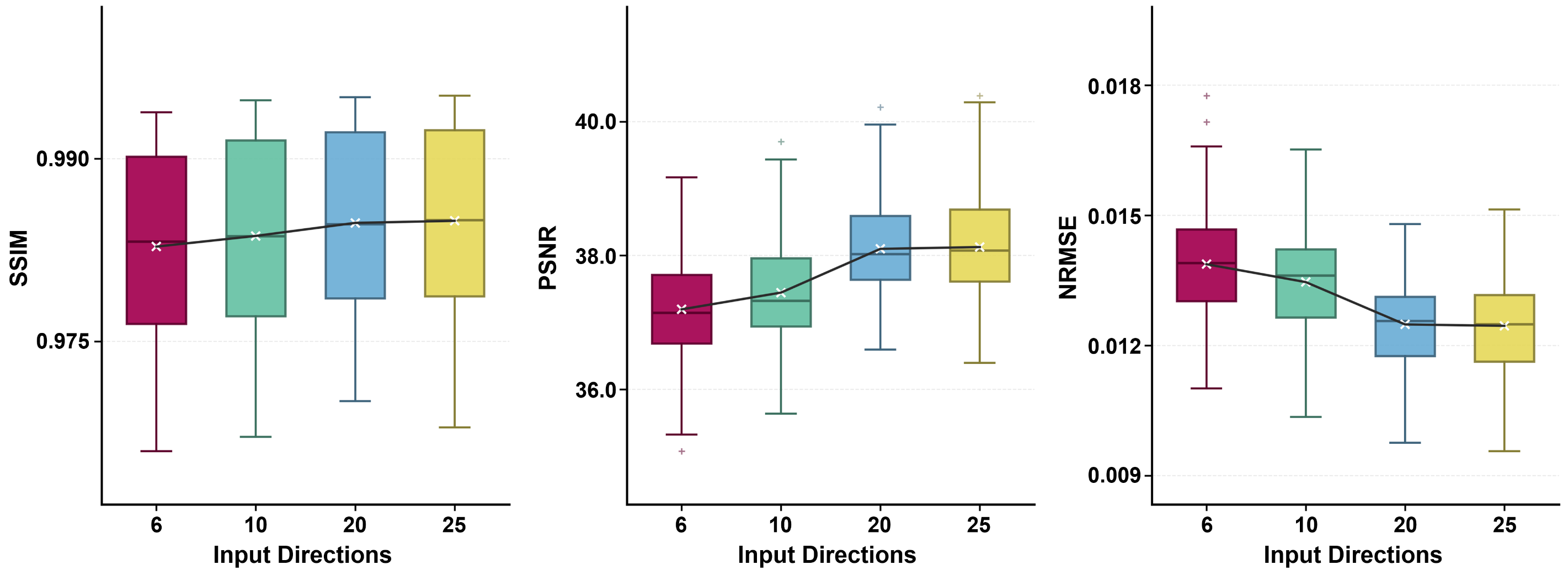}
\end{center}

\noindent\textbf{\textbf{Extended Data Fig. 2 | Effect of the number of input diffusion directions on DWI prediction performance.}}\par

Separate dMRI-F models were trained with 6, 10, 20, or 25 input DWI volumes to assess how the number of sampled diffusion directions affects model prediction accuracy. Each model was evaluated on the HCP-YA test set using the same number of input volumes as during training. Enhancement performance generally improved as the number of input directions increased. For all box plots, the center line denotes the median, the box spans the interquartile range, and the whiskers extend to the most extreme values within 1.5 times the interquartile range of the box. White crosses indicate means, connected by black lines to illustrate trends across input direction counts; plus signs denote outliers beyond the whiskers.
\end{figure}

\end{document}